\def\ARXIVVERSION{1}
\documentclass[letterpaper]{article} 
\ifdefined\ARXIVVERSION
\usepackage[preprint]{aaai2027}
\else
\usepackage[submission]{aaai2027} 
\fi
\usepackage[hyphens]{url} 
\usepackage{graphicx} 
\usepackage{natbib} 
\usepackage{caption} 
\usepackage{amsmath}
\usepackage{amssymb}
\usepackage{booktabs}
\newcommand{\method}{\textsc{CausalCache}}
\newcommand{\hgkv}{\textsc{HGKV}}
\newcommand{\base}{\pi_{0}}
\newcommand{\adapted}{\pi_{\mathrm{HG}}}

\title{CausalCache: Conditional High-Fidelity Restoration for Long-Horizon GUI Agents}
\ifdefined\ARXIVVERSION
\author{
Jiaxuan Luo\textsuperscript{\rm 1},
Zhanfeng Liao\textsuperscript{\rm 2},
Jiayao Teng\textsuperscript{\rm 1},
Yuan Wang\textsuperscript{\rm 2}
}
\affiliations{
\textsuperscript{\rm 1}Johns Hopkins University\\
\textsuperscript{\rm 2}Tsinghua University\\
jluo50@jhu.edu, zliao08@mails.tsinghua.edu.cn,\\
jteng68@jhu.edu, ywang80@mails.tsinghua.edu.cn
}
\else
\author{Anonymous Submission}
\affiliations{}
\fi

\begin{document}

\maketitle

\begin{abstract}
Long-horizon GUI agents can retain complete action histories as compact
text, but only a few historical screenshots fit in active context. We
formulate this as \emph{budgeted fidelity restoration}: every event
remains summarized, while a fixed budget $B$ determines which events
regain their archived screenshots. Recent-$B$ assigns all visual slots to
the latest events. \method{} instead scores the complete history and
swaps in an older event only when its predicted utility exceeds that of a
recent event. A history-gated key/value adapter modifies only restored
history-image tokens and is exactly bypassed when no history image is
active, preserving current-screen processing. The adapter and selector
are trained with matched-budget interventions on desktop trajectories and
evaluated zero-shot on mobile. On OSWorld-Verified, activating historical
screenshots improves success by about $13$ percentage points over
summary-only memory. Under the official $15$-step limit, \method{} and
Recent-$4$ are statistically indistinguishable; in a $30$-step
diagnostic, \method{} achieves $46.7\%$ success versus $42.4\%$
($+4.3$ points). Zero-shot on $117$ MobileWorld tasks, \method{} improves
over Recent-$4$ from $30.2\%$ to $36.8\%$. The gain is concentrated on a
pre-defined cross-app memory-candidate split ($30.6\%$ vs.\ $19.4\%$,
$+11.2$ points), while single-app controls show no detectable difference
($43.6\%$ vs.\ $42.4\%$). These results show that selecting \emph{which}
past events regain pixels is more effective than spending a fixed visual
budget entirely on recency.
\end{abstract}

\section{Introduction}

Autonomous agents that follow natural-language instructions and operate
graphical user interfaces promise broad practical value: they can automate
repetitive digital chores, carry out multi-application workflows on behalf
of their users, and extend what a single person can accomplish on desktop
and mobile devices \citep{xie2024osworld,rawles2025androidworld,
xu2026mobileagent35}. Delivering that value on long tasks turns memory into
a first-order design problem:
GUI agents accumulate screenshots, actions, text arguments, coordinates, and
interface changes over many decisions. A complete action trace persists
cheaply as compressed text, but keeping every screenshot active is costly and
distracting. Each past
event therefore has two policy-visible fidelities: a summary-only record and,
when needed, that same record with its archived screenshot reattached. The
memory-control question is not only which events remain recorded, but which
recorded events should be exposed again in high-fidelity pixels
\citep{lu2025guiodyssey,zeng2026mementogui,shi2026androtmem}.

Our central claim is that the object of control is \emph{event fidelity}, not
event inclusion. Existing controllers retain recent events, retrieve visually
similar screenshots, or learn salience from task supervision
\citep{zeng2026mementogui,shi2026androtmem,liu2026atmem}. Yet the value of
promoting a summarized event back to pixels depends on which images are already
active: without recent images, a distant screenshot may lack local grounding;
with a saturated recent window, another image may be redundant. Moreover, the
action policy must consume the reattached pixels. Our experiments show that a
strong frozen policy exhibits weak and unreliable \emph{average}
selectivity under matched replacement probes, although it can still
benefit from restored evidence on a subset of states; \hgkv{} sharpens
this content sensitivity and provides a more reliable utility signal for
training and deployment.

\begin{figure}[t]
\centering
\includegraphics[width=0.88\columnwidth]{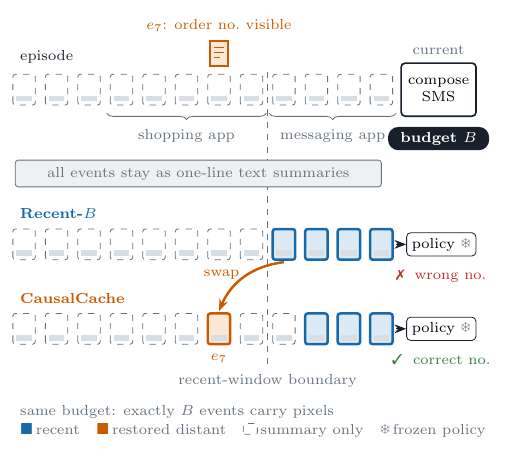}
\caption{\method{} on a real cross-app episode (schematic; drawn with $B{=}4$). The agent must compose an SMS containing the order number it
saw earlier in a shopping app. Every event persists as a one-line text
summary; a budget of $B$ events may additionally carry their archived
pixels. After the app switch the evidence frame $e_7$ lies outside the
recent window: Recent-$B$ spends every slot on the latest events and
types the wrong number, while \method{} swaps one recent slot for $e_7$
at the same budget and succeeds.}
\label{fig:qualitative-cart}
\end{figure}

\method{} accordingly treats GUI memory as \emph{conditional high-fidelity
restoration} (``causal'' denotes policy-specific effects of matched prompt
interventions, rather than structural causal discovery). Every event remains in the complete low-fidelity action trace,
and an archived screenshot is reattached only when that event is promoted into
the active visual context; no pixels are reconstructed from text. Under a
budget of $B$ high-fidelity history images, Recent-$B$ promotes the latest $B$
events. \method{} reallocates the same $B$ promotions over the complete trace,
with recent events as the default: a distant event enters only by displacing a
recent image whose conditional marginal utility it exceeds. Because every
compared allocation carries the same number of images, the contrast
isolates fidelity \emph{allocation} from visual capacity, and how many
non-recent events are actually promoted is measured as a result, never
preset. Figure~\ref{fig:qualitative-cart} illustrates this reallocation
on a real cross-application episode.

Learning to \emph{use} reattached pixels must not corrupt the base agent.
Full-layer adaptation simultaneously changes current-screen understanding,
action syntax, and grounding. \method{} instead trains a history-gated KV interface
(\hgkv) that touches only restored history-image tokens and is structurally
bypassed at zero budget, with a per-arm-anchored difference-of-differences
(DiD) objective in
which every arm is anchored to its own frozen score: uniform amplification of
any history cancels exactly, and only \emph{selective} use of target-relevant
content reduces the loss.

Our contributions are:
\begin{itemize}
    \item a conditional fidelity-restoration formulation of GUI memory:
    all events persist as compressed action records, while Recent-$B$ and
    \method{} allocate the same $B$ summary-plus-image promotions over that
    complete trace;
    \item a policy-preserving \hgkv{} interface trained on matched-budget
    replacement groups with per-arm-anchored DiD supervision, exact
    no-history parity, and pre-specified drift caps;
    \item a budget-aware restoration selector over the complete summarized
    trace, trained on desktop data only and evaluated zero-shot on mobile and
    desktop benchmarks.
\end{itemize}

\section{Related Work}

\paragraph{Visual GUI agents and benchmarks.}
WebShop, WebArena, Mind2Web, WorkArena, WorkArena++, VisualWebArena, and REAL
span grounded shopping, real websites, enterprise workflows, and visual web
interaction \citep{yao2022webshop,zhou2024webarena,deng2023mind2web,
drouin2024workarena,boisvert2024workarenaplus,koh2024visualwebarena,
garg2025real}. AITW, AndroidControl, AndroidWorld, OSWorld, and GUI-Odyssey
extend executable evaluation to mobile and desktop interfaces
\citep{rawles2023aitw,li2024androidcontrol,rawles2025androidworld,
xie2024osworld,lu2025guiodyssey}. Screenshot-native agents learn grounding and
actions directly from pixels \citep{hong2024cogagent,cheng2024seeclick,
lin2025showui,xu2026mobileagent35}. These works measure long-horizon
interaction; our question is which completed event should re-enter a bounded
prompt in pixels.
MobileWorld further stresses long-horizon, cross-application mobile workflows
under reproducible functional evaluation \citep{kong2026mobileworld}.

\paragraph{Agent memory and reusable experience.}
RAG retrieves external memory, ReAct interleaves reasoning with action, and
later agents preserve reflections, experience, or evolving long-term stores
\citep{lewis2020rag,yao2023react,shinn2023reflexion,zhao2024expel,
zhong2024memorybank,gutierrez2024hipporag}. ICAL retrieves multimodal programs
of thought \citep{sarch2024ical}. These systems write or retrieve artifacts
across interactions or tasks; \method{} instead holds a within-episode summary
trace fixed and controls whether an existing event is exposed as summary only
or summary plus image.

\paragraph{Long-horizon GUI memory.}
GUI-Odyssey resamples prior screenshots, MementoGUI learns memory operators,
and AndroTMem retrieves causally linked anchor states
\citep{lu2025guiodyssey,zeng2026mementogui,shi2026androtmem}. Their primary
variable is which item is retained or retrieved. \method{} keeps the compressed
trace fixed, controls event fidelity conditioned on the already-active visual
context, and trains the interface that consumes restored pixels.

\paragraph{Context and visual-token compression.}
Prompt compressors remove textual redundancy
\citep{jiang2023llmlingua,jiang2024longllmlingua,pan2024llmlingua2}; KV methods
retain, evict, select, or quantize cache states
\citep{xiao2024streamingllm,zhang2023h2o,liu2023scissorhands,
li2024snapkv,liu2024kivi,corallo2024finch}; and visual methods merge or prune
tokens within an image \citep{bolya2023tome,chen2024fastv,yang2025visionzip}.
Their unit is a token or KV position. \method{} instead preserves the summary
trace and allocates whole archived screenshots across events; \hgkv{} controls
only how promoted history-image tokens are consumed.

\paragraph{Interventional memory value.}
Recent work studies memory through controlled interventions or active state
influence \citep{srivastava2026cmi,liu2026atmem}. Our intervention is narrower
and executable: a selected event regains one archived post-action image while
the remaining prompt stays fixed, yielding a policy-specific behavioral
surrogate rather than an environment-level causal effect. Set models
parameterize coalition-conditioned scores
\citep{zaheer2017deepsets,lee2019settransformer}, whose marginals connect to
cooperative-game attribution \citep{shapley1953value}; we use controlled
restorations to train a deployable selector for one fixed GUI policy.

\section{Method}

\subsection{Problem Formulation}

At decision $t$, the agent receives instruction $g$, current observation
$x_t$, and history $H_t=\{e_1,\ldots,e_{t-1}\}$ with
$e_j=(x_j,a_j,x_{j+1})$, where each event has a compact low-fidelity
representation
$e_j^{\mathrm{low}}=\tilde e_j$, its one-line action summary, and an
available high-fidelity representation $e_j^{\mathrm{high}}=(r_j,v_j)$: the
official retained-turn form pairing the step's verbatim policy response
$r_j$ with its reattached archived post-action screenshot $v_j$. The
official protocol pairs every retained observation with the response that
produced it, so promotion and demotion move an event between these two
forms as a unit --- a fidelity allocation reassigns retained turns, not
images alone.

\paragraph{Fixed active-fidelity budget.}
Every prompt contains the goal $g$, the \emph{complete} textual action
summaries $\{\tilde e_1,\ldots,\tilde e_{t-1}\}$, and the current screenshot
$x_t$. An active visual-context budget $B$ limits how many summarized events
may be promoted by reattaching their archived pixels; it is not a limit on
cold archival storage. A restoration allocation is a set
$S\subseteq\{1,\ldots,t-1\}$ with $|S|=\min(B,\,t{-}1)$: the budget is
always filled, and reallocation is over \emph{which} events occupy it. Let
$Q(S)$ denote the policy's mean target-token log-likelihood of the successful
next action when events in $S$ are exposed in summary-plus-image form. The
standard baseline is $S_{\mathrm{recent}}=\mathrm{Recent}\text{-}B$, the $B$
most recent distinct events promoted to high fidelity. \method{} targets
\begin{equation}
S^{*}=\arg\max_{\substack{S\subseteq\{1,\ldots,t-1\}\\ |S|=\min(B,t-1)}} Q(S),
\qquad
Q(S^{*})\ge Q(S_{\mathrm{recent}}),
\label{eq:decision}
\end{equation}
so recency is one candidate fidelity allocation, not a separate modality.
We use \emph{causal} in a deliberately narrow, operational sense: for a
fixed policy, current observation, summarized trace, and active-image
budget, we intervene on the fidelity allocation and measure the resulting
change in policy behavior. The conditional marginal utility so defined is
policy- and prompt-specific; it is not an estimate of an
environment-level structural causal effect. The
realized replacement intensity $k=|S\setminus \mathrm{Recent}\text{-}B|$ is a
measured statistic, not a preset parameter. All comparisons at a given $B$
hold the image count, resolutions, underlying event trace, and prompt
structure fixed: every allocation renders exactly $|S|$ retained turns and
keeps every other event at summary fidelity; only which events occupy the
high-fidelity form differs. Cross-$B$ comparisons are
analysis, not the causal contrast. Given a partial allocation $S$, conditional
marginals
$\Delta_t(j\mid S)=Q(S\cup\{j\})-Q(S)$ drive selection, and a replacement is
justified only when the distant event's restoration marginal exceeds that of
the recent visual realization it displaces. Every formally reported set
utility reruns the complete restored set; we never report a sum of singleton
gains as the utility of a coalition.

\begin{figure*}[!t]
\centering
\includegraphics[width=0.76\textwidth]{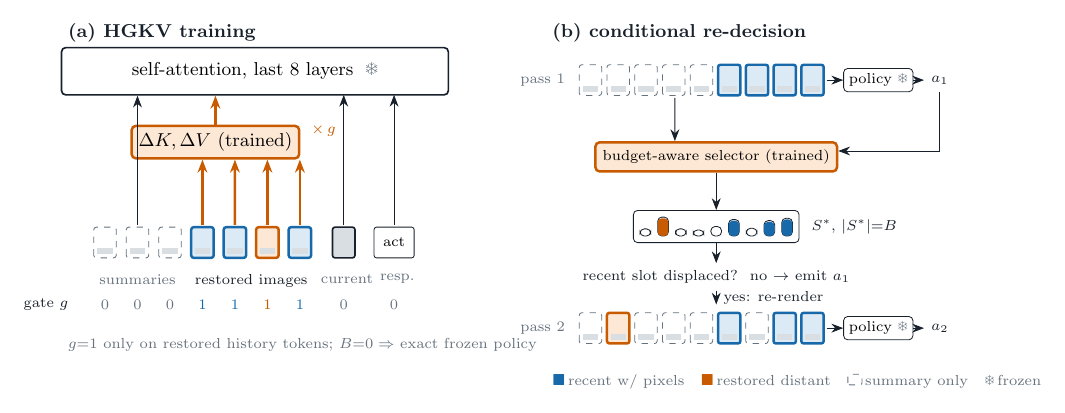}
\caption{The two learned components of \method{}.
(a)~The history-gated K/V interface (\hgkv): a low-rank
$\Delta K,\Delta V$ branch on the last eight frozen self-attention layers
is multiplied by a binary token gate $g$ that is $1$ only on restored
history-image tokens, so current-screen, text, and response tokens pass
through unchanged, and at $B{=}0$ the gate vanishes and the policy is
bitwise-identical to the frozen base.
(b)~Proposal-conditioned re-decision: pass~1 renders the
default Recent-$B$ allocation and yields a proposal $a_1$; a two-tower
scorer conditioned on per-event features with $a_1$ as witness and on
budget context scores the \emph{complete} summarized trace and returns the
exact-$B$ set $S^{*}$. If $S^{*}$ keeps the recent tail, $a_1$ is emitted
with no extra compute; only when a recent event is displaced is the prompt
re-rendered under $S^{*}$ and the \emph{same} frozen weights generate the
final action $a_2$. Event cards follow Figure~\ref{fig:qualitative-cart}: dashed gray
$=$ summary-only, blue $=$ recent with pixels, orange $=$ promoted
distant event; orange-outlined boxes are the trained modules; snowflakes
denote frozen components.}
\label{fig:overview}
\end{figure*}

\subsection{Policy-Preserving History-Gated KV Adapter}

Let $h_l$ be the input to layer $l$'s attention projections. In the last eight
language-model layers, \method{} adds rank-8, $\alpha{=}16$ low-rank residuals
\citep{hu2022lora} only to key and value projections:
\begin{equation}
\begin{aligned}
K_l &= W^K_l h_l + M_{\mathrm{hist}}\,\Delta W^K_l h_l,\\
V_l &= W^V_l h_l + M_{\mathrm{hist}}\,\Delta W^V_l h_l.
\end{aligned}
\end{equation}
$M_{\mathrm{hist}}$ is a token mask that is true only for image tokens
reattached to promoted historical events. It excludes instruction tokens,
event-summary text, the current observation image, and action tokens. The mask
is constructed fail-closed from the multimodal token-type sequence and
per-image patch geometry; any mismatch between declared images and encoded
image blocks aborts the forward pass. When no historical event is promoted,
the adapter context is absent and the projection hook returns the original
module output, so
$\adapted(\cdot\mid S{=}\varnothing)=\base(\cdot\mid S{=}\varnothing)$ up to bitwise equality
in our implementation.

\subsection{Per-$B$ Fidelity-Restoration Supervision}

Training states are decision points of successful desktop trajectories in
which the target action $a^*_t$ recurs earlier in the same trajectory under
full-action equivalence (type, arguments, and coordinates within tolerance);
the screenshot preceding the earlier occurrence is the target-specific
restoration candidate. For each state and each budget $B$ we construct
three matched-budget fidelity allocations with identical summaries, current
screen, and target---every prompt carries exactly $B$ history images:
\begin{itemize}
    \item \textbf{Recent}: $\mathrm{Recent}\text{-}B$, the $B$ most recent
    distinct events promoted to summary-plus-image form;
    \item \textbf{Relevant restore}: $\mathrm{Recent}\text{-}(B{-}1)+
    \{v_{j^{+}}\}$, the oldest recent visual slot replaced by the
    target-specific archived screenshot
    ($\mathrm{age}\geq B{+}2$, strictly older than the whole window);
    \item \textbf{Wrong restore}: the same slot replaced by an age-matched
    same-trajectory screenshot whose following action is \emph{not}
    equivalent to $a^*_t$.
\end{itemize}
Positives do not require the base action to be wrong: both information
addition and evidence amplification are admissible mechanisms.

\paragraph{Why single-slot replacement ($k{=}1$).}
Training teaches only the $k{=}1$ replacement interface: offline mining
finds simultaneous multi-frame demand rare, and a matched-budget probe
shows the first replacement is worth its slot while the paired increments
of a second and third are indistinguishable from zero or negative.
Larger-$k$ allocations remain expressible at deployment because the
selector composes exact-$B$ sets with recent fallback, so $k$ stays a
result statistic; mining details and the full probe are in the
supplement.

Let $\ell$ denote mean target log-likelihood, superscripted by adapter state
(active or frozen bypass). Each arm is anchored to \emph{its own} frozen
score:
\begin{equation}
\begin{aligned}
A_s&=\ell^{\mathrm{on}}_{\mathrm{relevant}}-\ell^{\mathrm{off}}_{\mathrm{relevant}},\qquad
A_r=\ell^{\mathrm{on}}_{\mathrm{recent}}-\ell^{\mathrm{off}}_{\mathrm{recent}},\\
A_n&=\ell^{\mathrm{on}}_{\mathrm{wrong}}-\ell^{\mathrm{off}}_{\mathrm{wrong}},
\end{aligned}
\end{equation}
and the DiD objective is
\begin{equation}
\begin{aligned}
\mathcal L ={}& [m-(A_s-A_r)]_+ + [m-A_s]_+ \\
&+ [m-(A_s-A_n)]_+ \\
&+ \lambda_{\mathrm{cap}}\bigl([|A_r|-\epsilon]_+ + [|A_n|-\epsilon]_+\bigr)
+ \lambda\lVert\Delta W\rVert_2^2,
\end{aligned}
\label{eq:did}
\end{equation}
with $m{=}0.01$, $\epsilon{=}0.02$, $\lambda_{\mathrm{cap}}{=}2$,
$\lambda{=}10^{-4}$, and no action cross-entropy. On the identity adapter
every increment is zero, so $A_s-A_r$ is exactly zero: the frozen policy's
own preference for target-relevant restorations cannot masquerade as adapter
skill, and uniformly amplifying all restored pixels buys nothing. The
dead-zone caps bound recent-window and wrong-history drift with gradients on
the same scale as the selection hinges. Losses and evaluation are stratified
by $B$; a pooled mean could hide a method that helps at $B{=}4$ but fails at
$B{=}1$.

\subsection{Budget-Aware Restoration Selector}

After the history interface is frozen, selector labels are true policy
utilities: singleton utilities for the \emph{entire} summarized event pool,
matched $Q(\mathrm{Recent}\text{-}B)$ anchors, and set utilities along
teacher paths, each fully re-scored as a complete prompt through the frozen
\hgkv{} policy. The selector is a set-conditional marginal scorer
$\widehat{\Delta}_t(j\mid S)$ trained on the implied marginals with a joint
regression and within-state ranking loss; it scores each candidate event
conditioned on the current action proposal, event recency, and the
already-selected set (the full feature list is in the supplement).
Deployment composes an exact-$B$ allocation with a small beam search
initialized at $\mathrm{Recent}\text{-}B$: a distant event enters only by
beating the predicted marginal of the recent realization it evicts,
realizing Eq.~\ref{eq:decision} as a relative comparison---declining every
replacement recovers Recent-$B$ exactly. Selector labels cover
$B\in\{1,2,4\}$; the realized replacement intensity $k$ is reported per
budget, never preset.

\paragraph{Reference actions.}
Part of the selector's input asks whether the action taken right after an
archived event matches a reference action. Training uses the gold
teacher-forced target, an offline oracle unavailable at deployment;
\method{} therefore re-grounds the reference in the policy's own tentative
action: the policy first acts on the Recent-$B$ prompt, and re-decides on
the re-allocated prompt only when a recent visual realization is
displaced. This proposal reference is statistically indistinguishable
from the gold oracle on held-out desktop groups, whereas weaker
references---no action signal, or instruction-only similarity---degrade
selection and over-replace; a single-pass variant reusing the previously
executed action removes the second call but transfers worse across
platforms (reference ladder and efficiency analysis in the supplement).

\section{Experiments}

\subsection{Experimental Setup}

\paragraph{Training Data and Model Selection.}

Policy adaptation and selector training use only desktop data: successful
AgentNet/OpenCUA Ubuntu trajectories (5{,}000 screened, 2{,}293 successful,
6{,}003 decision points) yield fixed-budget replacement groups at
$B{=}1/2/4$ of $969/764/470$ (2{,}203 in total; 1{,}774 training units, 211 development
groups, and 218 held-out test groups, trajectory-disjoint), complemented by a
high-precision OSWorld witness seed
mined from official successful trajectories. A further $160$ groups at
$B{=}8$ are constructed but deliberately excluded from training, so $B{=}8$
evaluates out-of-training budget extrapolation. Checkpoints and thresholds
are selected only on the desktop development split under pre-specified
gates: positive DiD selection with a positive bootstrap lower bound, and
drift caps $|A_r|,|A_n|<\epsilon$. No mobile benchmark data touches training
or selection. All adapter training, offline scoring, and closed-loop serving
run on NVIDIA H200 Tensor Core GPUs.

\paragraph{The Fixed-Budget Fidelity-Restoration Design.}

Figure~\ref{fig:overview} shows the two learned components and the
same-budget re-decision that defines the core comparison.
Every compared allocation carries exactly $B$ history images with
identical resolutions, summaries, and prompt structure, so the contrast
isolates \emph{which summarized events are promoted to high fidelity}.
$B{=}4$ is the primary deployment-like setting; the realized number of
non-recent promotions $k$ is measured rather than preset.

\paragraph{Policies and Fidelity-Allocation Baselines.}

The frozen policy is GUI-Owl-1.5-8B-Instruct \citep{xu2026mobileagent35}.
Under identical data, steps, and cadence we compare the frozen policy, an
ungated KV control (the same last-eight layers, K/V targets, rank, and
alpha as \hgkv{} but active on all tokens), and \hgkv{} itself
(token-gated, structurally bypassed at $B{=}0$).
Fidelity-allocation baselines are Recent-$B$, Random, OCR/RGB similarity, and
the learned budget-aware restoration selector (exact-$B$, recent fallback); a
beam oracle over the candidate pool bounds headroom. Every method receives the
same summaries and the same active image budget.

\paragraph{Zero-Shot Benchmarks and Metrics.}

The policy interface and selector, trained on desktop data only, are evaluated
(i) on held-out desktop decision groups (offline, per-$B$), (ii) zero-shot on
an uncontaminated OSWorld task roster \citep{xie2024osworld} with official
evaluator scores (tasks whose trajectories seeded the witness corpus are
excluded from zero-shot claims and reported separately as in-domain
diagnostics), and (iii) zero-shot on a cross-app mobile memory benchmark of
117 runnable tasks (62 cross-app memory candidates vs.\ 55 single-app
controls, split mechanically by the benchmark's task metadata),
which contributes no data to any training or selection decision
\citep{kong2026mobileworld}.
Offline metrics are teacher-forced log-likelihood margins and full-action
equivalence under an argument and coordinate tolerance (sensitivity settings
in the supplement); closed-loop metrics are paired task success with
cluster bootstrap, step counts, and wall time. All arms decode greedily,
so every action is a deterministic function of its rendered prompt, and
the conditional second pass always replaces the proposal rather than
selecting between samples---it cannot act as best-of-$n$.

\subsection{Results}

\begin{table}[t]
\centering
{\small
\setlength{\tabcolsep}{3.2pt}
\begin{tabular}{@{}lrrr@{}}
\toprule
Benchmark / stratum & Recent-$B$ & \method{} & $\Delta$ \\
\midrule
MobileWorld full ($117$) & $30.2$ & $36.8$ & $+6.6^{*}$ \\
\hspace{7pt}memory-critical ($62$) & $19.4$ & $30.6$ & $+11.2^{*}$ \\
\hspace{7pt}single-app control ($55$) & $42.4$ & $43.6$ & $+1.2$ \\
\hspace{7pt}split$\times$method interact. & --- & --- & $+10.1$ \\
OSWorld-Verified ($361$) & $33.0$ & $33.3$ & $+0.3$ \\
\hspace{7pt}$30$ max steps & $42.4$ & $46.7$ & $+4.3^{*}$ \\
\bottomrule
\end{tabular}
}
\caption{Primary closed-loop contrasts at the deployment budget $B{=}4$
(task-paired bootstrap; MobileWorld three-run means; the OSWorld row uses the official $15$-step cap). $^{*}$~the $95\%$
confidence interval strictly excludes zero (full CI tables in the
supplement; the interaction interval accompanies them). The gain
concentrates exactly on the construction-defined memory-critical stratum,
with no detectable effect on controls; on desktop the allocations tie at
the official $15$-step cap and separate when it is doubled to $30$.}
\label{tab:contrast}
\end{table}

\begin{table*}[!t]
\centering
{\small
\setlength{\tabcolsep}{5.0pt}
\begin{tabular*}{\textwidth}{@{\extracolsep{\fill}}lrrrr@{}}
\toprule
Method & OSWorld-Verified ($361$) & MobileWorld ($117$) &
MW memory-critical ($62$) & MW control ($55$) \\
\midrule
Frozen, summary-only ($B{=}0$) & $19.9$ & $28.2$ & $21.0$ & $36.4$ \\
Frozen $+$ Recent-$B$ & $32.7$ & $29.9$ & $22.6$ & $38.2$ \\
\hgkv{} $+$ Recent-$B$ & $33.0$ & $30.2$ & $19.4$ & $42.4$ \\
Frozen $+$ selector & $32.7$ & $33.9$ & $28.0$ & $40.6$ \\
\method{} (\hgkv{} $+$ selector) & $\mathbf{33.3}$ & $\mathbf{36.8}$ &
$\mathbf{30.6}$ & $\mathbf{43.6}$ \\
\hspace{7pt}frozen-taught selector variant & --- & $33.3$ & $25.8$ & $41.8$ \\
\bottomrule
\end{tabular*}
}
\caption{Closed-loop success rates (\%) at the primary deployment budget
$B{=}4$; the offline budget sweep is in the supplement.
OSWorld-Verified: official no-GDrive roster and official $15$-step budget,
one run per arm (a $30$-step extended-horizon rerun of both closed-loop
arms is reported in the text). MobileWorld: zero-shot means over three
runs for all multi-run arms; the memory-critical column restricts to the $62$
construction-defined cross-app tasks, where the allocation contrast is
largest, and the control column to the $55$ single-app tasks, where no
restoration method should help. \method{} deploys the \hgkv{} adapter
plus the budget-aware selector; Frozen $+$ selector runs the same
selector on the fully frozen policy, and the frozen-taught variant
additionally trains the selector on frozen-policy-scored labels.
High-fidelity arms beat summary-only memory on desktop but remain
pairwise indistinguishable there. Full paired statistics and per-round
results are in the supplementary document.}
\label{tab:closed-loop}
\end{table*}

\paragraph{Selectivity Is Not Inherited from the Frozen Policy.}

On the desktop development split, the frozen policy shows no reliable
preference for a task-relevant archived screenshot over the informative
recent frame it would displace (recent-baseline construction details in
the supplement): to the frozen policy the two are worth about the same,
so profitable fidelity reallocation must be learned. Under the official
multi-turn protocol the frozen replacement effect decays with the window
and reverses at large $B$ (supplement), while the learned
interface keeps a positive selection effect across the budget axis.

\paragraph{Selective Use Within a Drift Envelope.}

Table~\ref{tab:did-gate} reports the pre-specified desktop DiD gate at
$B{=}1$. Adapters are trained for matched steps; checkpoints follow the
frozen selection rule. \hgkv{} passes every gate: its selected checkpoint
learns a positive DiD selection effect against the redundant-duplicate
recent control, a re-scoring probe confirms the effect against the
informative true-previous-frame control, and recent-only and wrong-history
drift stay well inside the cap $\epsilon{=}0.02$; its no-history path is
bitwise identical to the frozen policy under randomized nonzero adapter
weights. The layer- and rank-matched ungated control also learns
selectivity, confirming that the last-eight K/V position carries signal,
but token gating adds a significant paired per-group advantage while
holding drift further from the caps. Gating is what buys selectivity
\emph{inside} the envelope.

\begin{table}[t]
\centering
{\small
\setlength{\tabcolsep}{3.6pt}
\begin{tabular}{@{}lrrr@{}}
\toprule
Adapter & DiD select & $|A_r|$ & Wrong drift \\
\midrule
\textbf{\hgkv{}} & $\mathbf{+0.0224}$ & $0.0088$ & $0.0085$ \\
Ungated KV & $+0.0177$ & $0.0086$ & $0.0071$ \\
\bottomrule
\end{tabular}
}
\caption{Desktop DiD evaluation on 94 development groups ($B{=}1$).
DiD select is the target-relevant conditional margin ($A_s-A_r$); both
adapters hold recent-window ($|A_r|$) and wrong-history drift within the
cap $0.02$ (intervals and all budgets in the supplement). The \hgkv{}
zero-budget path is exactly the frozen policy.}
\label{tab:did-gate}
\end{table}

\paragraph{Fidelity Reallocation and Zero-Shot Transfer.}

The offline budget axis shows the same picture. Across the trained budgets
$B\in\{1,2,4\}$, \hgkv{} maintains positive selectivity---per-$B$ DiD
gates all pass with drift within the pre-specified caps---and the learned
allocation improves over Recent-$B$ at matched image counts, while keeping
the full recent window in $19\%$ of $B{=}4$ states. The ungated control
again passes but with uniformly smaller selection effects, drifting
toward the cap at the largest budget. Untrained-budget extrapolation
($B{=}8$) and the full per-budget tables are in the supplement.

\paragraph{Mobile closed loop (zero-shot).}
The 117-task mobile benchmark contributes no training or selection
signal and is our primary test of \emph{allocation}: its roster carries
a memory-critical split deterministically partitioned via the
benchmark's upstream metadata (multi-app vs.\ single-app), with zero
manual filtering.
The full-roster aggregates are secondary: over three runs at the primary
budget $B{=}4$, \method{} attains the highest overall success rate,
$36.8\%$, against $30.2\%$ for \hgkv{}+Recent-$B$ and $28.2\%$ for the
frozen summary-only baseline (Table~\ref{tab:closed-loop}), with the same
sign in every round. The decomposition is clean:
frozen Recent-$B$, \hgkv{}+Recent-$B$, and the summary-only baseline are
statistically indistinguishable ($29.9/30.2/28.2\%$) --- the behavioral
counterpart of the drift cap --- so the gain is carried by \emph{which
summarized events are promoted}, not by the adapter or by recency alone.
Running the same selector on the fully frozen policy retains most of the
margin ($33.9\%$), and re-teaching it from frozen-policy-scored labels
retains less ($33.3\%$): the adapter contributes both at deployment and
as the utility teacher. A
recent-dose sweep of the frozen policy makes the last point explicit:
across $B{=}0$--$4$ success is non-monotone with every paired contrast
against $B{=}0$ crossing zero; promoting additional recent events alone
buys nothing, so fidelity \emph{allocation} rather than active visual
budget \emph{size} is the operative variable.
\begin{figure}[t]
\centering
\includegraphics[width=0.82\columnwidth]{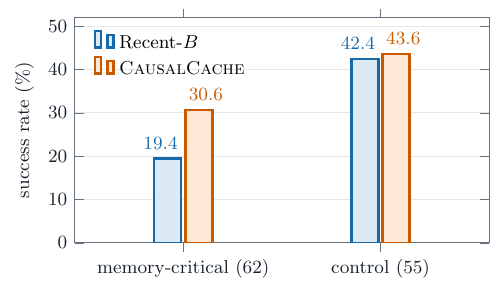}
\caption{The core contrast at the primary budget: on construction-defined
memory-critical tasks \method{} gains $+11.2$pp over the same-budget recent
allocation, while on matched single-app controls no effect is detectable
($+1.2$pp); the split-by-method interaction is $+10.1$pp (paired
intervals in the supplement).}
\label{fig:split-bars}
\end{figure}

The primary closed-loop result is that split: cross-app tasks are
constructed so that information from an earlier application is needed
after the switch. The rule uses no model outcome, and candidates that do
not truly require distant evidence only dilute the stratum toward zero,
making the contrast conservative (rule text and manifest hashes in the
supplement). On the
memory-critical stratum the same-budget contrast is decisive
(Figure~\ref{fig:split-bars}):
\method{} $30.6\%$ vs.\ Recent-$B$ $19.4\%$, a $+11.2$pp gain; on the
controls no effect is detectable ($43.6\%$ vs.\ $42.4\%$); and the
split-by-method interaction is $+10.1$pp (intervals
in the supplement). The value of reallocation is conditional on distant visual
dependency: the modest full-roster mean is composition, not absence of
effect, since $47\%$ of the roster is control tasks on which no high-fidelity
restoration method should help.

As a complementary outcome-defined analysis, we call tasks that the frozen
summary-only policy cannot complete within the official budget
\emph{memory-demanding}, and report
$\Delta_{\text{long}}=P(\text{success}\mid\text{\method},\text{long})-P(\text{success}\mid\text{baseline},\text{long})$
on that stratum. Across $95$ such mobile tasks, \method{} gains
$7.0$pp over summary-only memory, while adding nothing on the $22$ tasks
the baseline already solves. These post-hoc strata are secondary to the construction-defined
memory-critical split.

\paragraph{Desktop closed loop.}
On OSWorld-Verified (361 tasks, official no-GDrive roster and
$15$-step budget), all high-fidelity arms reach $32.7$--$33.3\%$
(\method{} $33.3\%$), each
about $+13$pp over summary-only memory ($19.9\%$, paired $95\%$ CIs
excluding zero), while same-budget allocations remain
indistinguishable. The official step budget itself binds, however:
$352$ of $361$ episodes exhaust the $15$-step cap. Rerunning both arms
with the cap doubled to $30$ steps separates the allocations: Recent-$4$
rises to $42.4\%$ ($+9.5$pp from the extra horizon) while \method{}
rises to $46.7\%$ ($+13.4$pp), converting the extended horizon into
$+3.9$pp more in-domain success and a $+4.3$pp margin at $30$ steps
versus $+0.3$ at $15$. Under the official cap OSWorld therefore
supports high-fidelity \emph{availability}; conditional
\emph{allocation} shows on the horizon-extended desktop run and is
tested zero-shot by the mobile memory-critical split. \hgkv{} again shows no detectable behavioral difference on matched
Recent-$B$ allocations, mirroring the mobile finding. \method{} adds a
median of eight policy calls per task, about $7\%$ of median end-to-end
task time.

\subsection{Ablations and Analyses}

\looseness=-1 Policy-side per-$B$ gates and the frozen recent-dose sweep appear in Results.
Table~\ref{tab:chooser} brackets the learned selector between allocation
rules that share its budget and rendering exactly. Random exact-$B$ and
RGB-similarity restoration are indistinguishable from Recent-$B$: sparse
exposure of archived pixels, or visual similarity to the current screen, is
not utility. Inverting the learned score is significantly harmful at both
budgets---a specificity control showing the scorer ranks real signal
symmetrically rather than exploiting a rendering artifact. Retraining
without the recency-membership features returns the selector to the noise
band, so budget-context features carry substantial weight.

\begin{table}[t]
\centering
{\small
\setlength{\tabcolsep}{4.0pt}
\begin{tabular}{@{}lcc@{}}
\toprule
Allocation rule & $B{=}2$ & $B{=}4$ \\
\midrule
Learned scorer & $+.030$ & $+.017$ \\
Random exact-$B$ & $+.006$ & $+.008$ \\
RGB similarity & $+.006$ & $+.007$ \\
Inverted score & $-.016$ & $-.009$ \\
No recency feats & $+.012$ & $+.008$ \\
\bottomrule
\end{tabular}
}
\caption{Chooser ladder: log-likelihood margin over Recent-$B$ at matched
budgets with fresh full-set policy scoring. Only the learned ranking and
its inversion separate from zero (episode-clustered intervals in the
supplement).}
\label{tab:chooser}
\end{table}

\looseness=-1 The budget is spent adaptively: at $B{=}4$ the learned scorer keeps the
entire recent window on $19\%$ of states and replaces all of it on
$26\%$---a state-dependent $k$ profile that neither random allocation nor
RGB similarity reproduces. Although training supervises only the
$k{=}1$ interface, these multi-slot compositions sustain the closed-loop
gains of Table~\ref{tab:closed-loop}: the token-gated adaptation applies
per restored frame and generalizes across replacement multiplicity.

\begin{figure}[t]
\centering
\includegraphics[width=0.76\columnwidth]{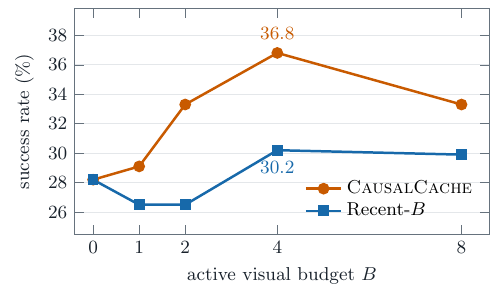}
\caption{Closed-loop MobileWorld success along the budget axis (full
roster). $B{=}0/4$: three-run means; other budgets single rounds
($B{=}1/2$ references: frozen dose curve; \hgkv{} neutral on matched
recent). Pooled over the sweep: $+4.9$pp (task-paired bootstrap;
interval and full per-budget table in the supplement).}
\label{fig:budget-curve}
\end{figure}

\paragraph{Closed-loop budget sweep.}
Figure~\ref{fig:budget-curve} extends the closed loop along the deployment budget
axis with single-round mobile campaigns at $B\in\{1,2,8\}$, sharing every
frozen weight with the primary setting. The allocation gain is positive at
every budget and, pooled over $B\in\{1,2,4,8\}$ with equal budget weight,
reaches $+4.9$pp (task-paired bootstrap; interval in the supplement), while
single-budget cells remain trend evidence. On the memory-critical subset
the margin is largest at the trained budgets and narrows at the untrained
$B{=}8$, where the eight-frame recent window already reaches back far
enough to cover part of the split's evidence---further support that the
operative variable is where the evidence sits relative to the window.

\paragraph{What the restored slots must carry.}
Two controls keep selection, budget, and the two-pass interface intact
but replace each restored frame's pixels with text: the turn's verbatim
recorded response, or an OCR transcription of the same screenshot (both
run on the selector$+$frozen deployment, whose pixel version scores
$28.0\%$ on this split). On the memory-critical split verbatim text
collapses to the Recent-$4$ level ($18.0\%$ vs.\ $19.4\%$; $-10.4$pp
below pixels, task-paired $p{=}0.006$), ruling out a generic
``reminder'' effect; OCR recovers part of the level ($24.2\%$) but its
$+4.8$pp margin over Recent-$4$ does not separate from zero, while pixel
restoration's $+8.6$pp does. Full tables in the supplement.

\paragraph{Deployment cost of the second pass.}
The conditional second pass fires on
$81$--$87\%$ of audited mobile steps; the selector costs under $40$\,ms. Instrumented across budgets, the added model time per
step is $2.0/2.3/2.9$\,s at $B{=}1/2/4$---$8.1/8.7/10.5\%$ of the measured
per-step wall time---and falls to roughly $5\%$ at $B{=}8$, where the
second pass fires on only $39\%$ of steps. On desktop the median
end-to-end task overhead is $7\%$. The re-decision's real-time cost thus
peaks near one-tenth of wall time at the primary budget; complete
intervals are in the supplement.

\section{Limitations and Conclusion}
\enlargethispage{2\baselineskip}

\looseness=-1
The evidence assembles into three steps. First, the problem is fidelity
allocation under a fixed visual budget: every event survives as text and
only $B$ regain pixels.
Second, a frozen policy does not reliably separate a useful distant
screenshot from the recent frame it would displace; \hgkv{} learns to,
inside a drift envelope that leaves matched-recent behavior untouched.
Third, the learned selector reallocates profitably exactly where distant
visual dependency exists: memory-critical tasks improve, controls show no
detectable change.
We control an active context budget, not archive capacity, and
\method{} reattaches rather than generates pixels. We claim only that
some summarized events have greater conditional marginal utility than
the recent realization they displace; pretraining contamination remains
possible. In one sentence: \method{} learns
which summarized GUI events deserve to be seen again.

\clearpage
\bibliography{references}

\end{document}